\def\year{2020}\relax
\documentclass[letterpaper]{article} % DO NOT CHANGE THIS
\usepackage{aaai20}  % DO NOT CHANGE THIS
\usepackage{times}  % DO NOT CHANGE THIS
\usepackage{helvet} % DO NOT CHANGE THIS
\usepackage{courier}  % DO NOT CHANGE THIS
\usepackage[hyphens]{url}  % DO NOT CHANGE THIS
\usepackage{graphicx} % DO NOT CHANGE THIS
\usepackage{graphicx}  % DO NOT CHANGE THIS
\usepackage{tikz}

\usepackage{amsmath}
\usepackage{array}
\usepackage{subcaption}
\usepackage{multirow}

\title{Hearing Lips: Improving Lip Reading by Distilling Speech Recognizers}

\author{Ya Zhao,\textsuperscript{\rm 1} Rui Xu,\textsuperscript{\rm 1} Xinchao Wang,\textsuperscript{\rm 2} Peng Hou,\textsuperscript{\rm 3} Haihong Tang,\textsuperscript{\rm 3} Mingli Song\textsuperscript{1}\\ % All authors must be in the same font size and format. Use \Large and \textbf to achieve this result when breaking a line
\textsuperscript{\rm 1}Zhejiang University, 
\textsuperscript{\rm 2}Stevens Institute of Technology, 
\textsuperscript{\rm 3}Alibaba Group
\\ 
\{yazhao, ruixu, brooksong\}@zju.edu.cn,
xinchao.wang@stevens.edu,\\
houpeng.hp@alibaba-inc.com, piaoxue@taobao.com
}

\begin{document}
\maketitle

\begin{abstract}
Lip reading has witnessed unparalleled development in recent years 
thanks to deep learning and the availability of large-scale datasets. 
Despite the encouraging results achieved, the performance of lip reading, unfortunately, remains inferior to the one of its counterpart speech recognition, 
due to the ambiguous 
nature of its actuations that makes it challenging to extract discriminant features 
from the lip movement videos.
In this paper, we propose
a new method, 
termed as Lip by Speech~(LIBS),
of which the goal is to strengthen
lip reading by learning from speech recognizers. 
The rationale behind our approach is that
the features extracted from
speech recognizers may provide 
complementary and discriminant clues,
which are formidable to be obtained
from the subtle movements of the lips,
and consequently facilitate the training of lip readers.
This is achieved, specifically, 
by distilling multi-granularity knowledge
from speech recognizers to lip readers. 
To conduct this cross-modal knowledge distillation, 
we utilize an efficacious alignment scheme to handle the inconsistent lengths
of the audios and videos, as well as an innovative filtering strategy to refine the speech recognizer's prediction.
The proposed method achieves the new state-of-the-art performance on the CMLR and LRS2 datasets, outperforming the baseline by a margin of 7.66\% and 2.75\% in character error rate, respectively. 
\end{abstract}

\section{Introduction}
Lip reading, also known as visual speech recognition, 
aims at predicting the sentence being spoken, 
given a muted video of a talking face. 
Thanks to the recent development of deep 
learning and the availability of big data for training, 
lip reading has made unprecedented progress with much performance enhancement~\cite{assael2016lipnet,chung2017lipWild,zhao2019cascadelipreading}. 

In spite of the promising accomplishments, the performance of the
video-based lip reading remains considerably lower than its counterpart, 
the audio-based speech recognition,
for which the goal is also to decode the spoken text 
and therefore can be treated as 
a heterogeneous modality sharing the same underlying 
distribution as lip reading. 
Given the same amount of training data and model architecture,
the performance discrepancy
is as large as 10.4\% vs. 39.5\% in terms of 
character error rate
for speech recognition and lip reading, respectively~\cite{chung2017lipWild}.
This is due to the intrinsically ambiguous nature of lip actuations: 
several seemingly-identical lip movements may
produce different words, making it highly challenging 
to extract discriminant features from the video of interest
and to further dependably predict the text output.

In this paper, we propose a novel scheme, Lip by Speech~(LIBS),
that utilizes speech recognition, for which the performances 
are in most cases gratifying, to facilitate the training of 
the more challenging lip reading.  We assume a pre-trained 
speech recognizer is given, and attempt to distill knowledge 
concealed in the speech recognizer to the target lip reader 
to be trained.

The rationale for exploiting 
knowledge distillation~\cite{hinton2015distilling}
for this task lies in that,
acoustic speech signals embody
information complementary to 
that of the visual ones.
For example, utterances with subtle movements, 
which are challenging to be distinguished visually,
are in most cases handy to be recognized 
acoustically~\cite{wolff1994lipreading}. 
By imitating the acoustic speech features 
extracted by the speech recognizer, 
the lip reader is expected to enhance its 
capability to extract discriminant visual features.
To this end,  LIBS is designed to distill knowledge 
at multiple temporal scales 
including sequence-level, context-level, and frame-level,
so as to encode the multi-granularity semantics 
from the input sequence. 

Nevertheless, distilling knowledge from a heterogeneous 
modality, in this case the audio sequence, confronts two major challenges. 
The first lies in the fact that, the two modalities may feature 
different sampling rates and are thus asynchronous, while the second 
concerns the imperfect speech-recognition predictions. 
To this end, 
we employ a cross-modal alignment strategy to
synchronize the audio and video data by 
finding the correspondence between them,
so as to conduct the fine-grained knowledge distillation 
from audio features to visual ones.
To enhance the speech predictions, on the other hand, 
we introduce a filtering technique to refine the distilled features,
so that useful features can be filtered for knowledge distillation.

Experimental results on two large-scale lip reading datasets,
CMLR \cite{zhao2019cascadelipreading} and LRS2 \cite{afouras2018deep}, show that the proposed approach outperforms the  
state of the art. 
We achieve a character error rate of 31.27\%,
a 7.66\% enhancement over the baseline on the CMLR dataset,
and one of 45.53\% with 2.75\% improvement on LRS2. 
It is noteworthy that when the amount of training data shrinks, the proposed approach tends to yield an even greater performance gain. For example, when only 
20\% of the training samples are used, the performance against the baseline has an 9.63\% boost on the CMLR dataset. 

Our contribution is therefore an innovative and effective approach to 
enhancing the training of lip readers, 
achieved by distilling multi-granularity knowledge
from speech recognizers. 
This is to our best knowledge the first attempt along this line and, 
unlike existing feature-level knowledge distillation methods
that work on Convolutional Neural Networks~\cite{romero2014fitnets,gupta2016cross,hou2019learning},
our strategy handles  Recurrent Neural Networks.
Experiments on several datasets show that 
the proposed method leads to the new state of the art.

\section{Related Work}
\subsection{Lip Reading}
\cite{assael2016lipnet} proposes the first deep learning-based, end-to-end sentence-level lipreading model. It applies a spatiotemporal CNN with Gated Recurrent Unit (GRU) \cite{cho2014learning} and Connectionist Temporal Classification
(CTC) \cite{graves2006connectionist}.
\cite{chung2017lipWild} introduces the WLAS network utilizing a novel dual attention mechanism that can operate over visual input only, audio input only, or both.  
\cite{afouras2018deep} presents a seq2seq and a CTC architecture based on self-attention transformer models, and are pre-trained on a non-publicly available dataset. 
\cite{shillingford2018large} designs a lipreading system that uses a network to output phoneme distributions and is trained with CTC loss, followed by finite state transducers with language model to convert the phoneme distributions into word sequences. 
In \cite{zhao2019cascadelipreading}, a cascade sequence-to-sequence architecture (CSSMCM) is proposed for Chinese Mandarin lip reading. 
CSSMCM explicitly models tones when predicting characters. 

\subsection{Speech Recognition}
Sequence-to-sequence models are gaining popularity in the automatic speech recognition (ASR) community, since it folds separate components of a conventional ASR system into a single neural network. 
\cite{chorowski2014end} combines sequence-to-sequence with attention mechanism to decide which input frames be used to generate the next output element.
\cite{chan2016listen} proposes a pyramid structure in the encoder, which reduces the number of time steps that the attention model has to extract relevant information from. 

\subsection{Knowledge Distillation}
Knowledge distillation is originally introduced for a smaller student network to perform better by learning from a larger teacher network \cite{hinton2015distilling}. 
The teacher network has previously been trained, and the parameters of the student network are going to be estimated. 
In \cite{romero2014fitnets}, the knowledge distillation idea is applied in image classification, where a student network is required to learn the intermediate output of a teacher network. 
In \cite{gupta2016cross}, knowledge distillation is used to teach a new CNN for a new image modality (like depth images), by teaching the network to reproduce the mid-level semantic representations learned from a well-labeled image modality. 
\cite{kim2016sequence} propose a sequence-level knowledge distillation method for neural machine translation at the output level. 
Different from these work, we perform feature-level knowledge distillation on Recurrent Neural Networks.

\section{Background}
\begin{figure*}[ht]
\centering
\includegraphics[width=1.8\columnwidth]{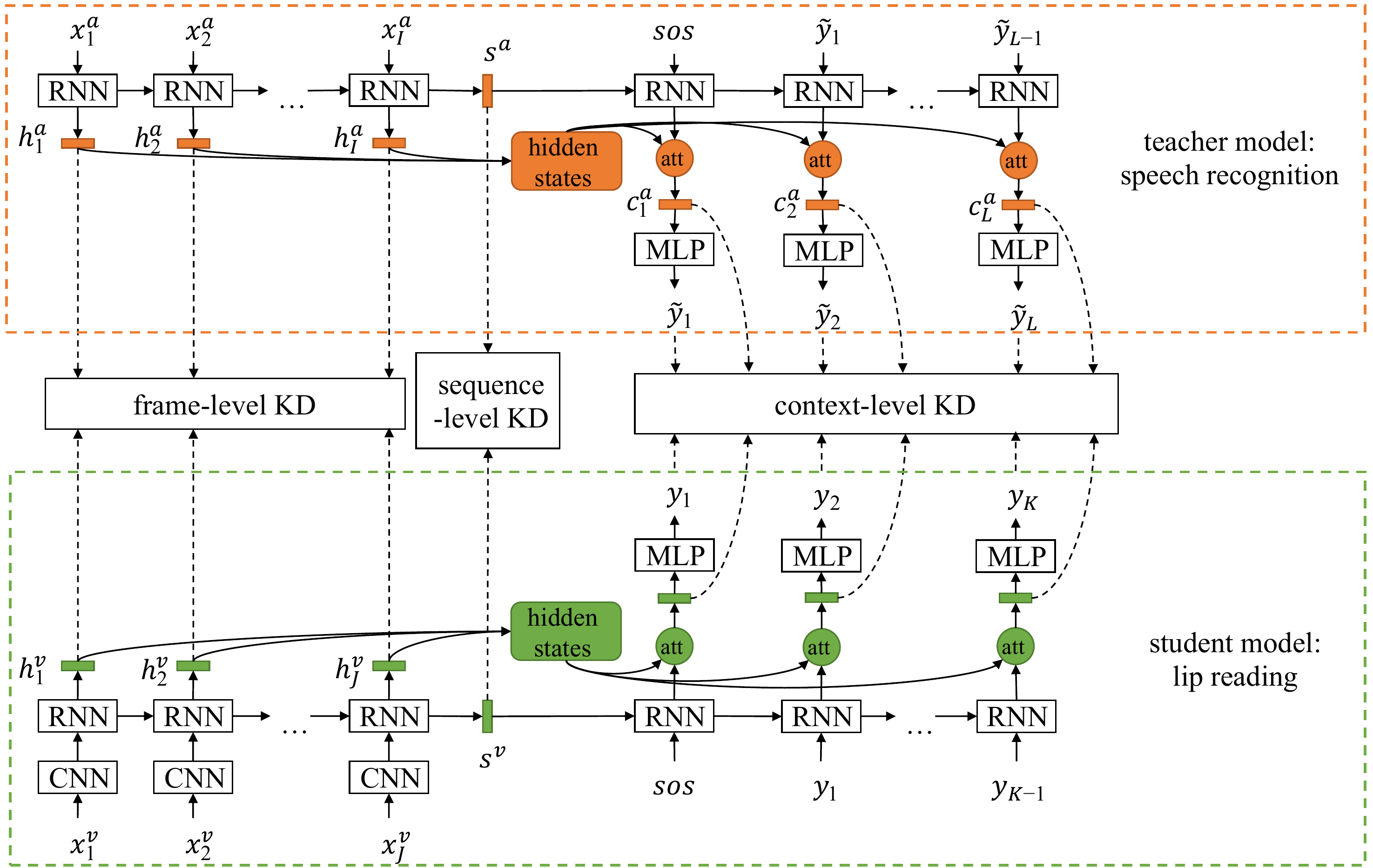}
\caption{The framework of LIBS. 
The student network deals with lip reading, 
and the teacher handles speech recognition. 
Knowledge is distilled at sequence-, context-, and frame-level to  
enable the features of multi-granularity to be
transferred from teacher network to student. KD is short for knowledge distillation.}
\label{fig:network}
\end{figure*}

Here we briefly review the attention-based sequence-to-sequence model \cite{bahdanau2015neural}.

Let $\textbf{x} = [x_1, ..., x_I]$, $\textbf{y} = [y_1, ..., y_K]$ be the input and target sequence with a length of $I$ and $K$ respectively. 
Sequence-to-sequence model parameterizes the probability $p(\textbf{y}|\textbf{x})$ with an encoder neural network and a decoder neural network. 
The encoder transforms the input sequence $x_1, ..., x_I$ into a sequence of hidden state $h^x_1, ..., h^x_I$ and produces the fixed-dimensional state vector $s^x$, which contains the semantic meaning of the input sequence. 
We also called $s^x$ the \textit{sequence vector} in this paper.
\begin{equation}
    h^x_i = {\rm RNN}(x_i, h^x_{i-1}),
\end{equation}
\begin{equation}
    s^x = h^x_I.
\end{equation}

The decoder computes the probability of the target sequence conditioned on the outputs of the encoder. Specifically, given the input sequence and previously generated target sequence $y_{<k}$, the conditional probability of generating the target $y_k$ at timestep $k$ is decided by:
\begin{equation}
\begin{split}
    p(y_k|y_{<k}, \textbf{x}) = g(y_{k-1}, h^d_k, c^x_k), \\
    h^d_k = {\rm RNN}(h^d_{k-1}, y_{k-1}, c^x_k),
\end{split}
\end{equation}
where $g$ is the softmax function, $h^d_k$ is the hidden state of decoder RNN at timestep $k$, and $c^x_k$ is the context vector calculated by an attention mechanism. 
Attention mechanism allows the decoder to attend to different parts of the input sequence at each step of output generation. 

Concretely, the context vector is calculated by weighting each encoder hidden state $h^x_i$ according to the similarity distribution $\alpha_{k}$:
\begin{equation}
    c^x_k = \sum^I_{i=1}\alpha_{ki}h^x_i, 
\end{equation}
The similarity distribution $\alpha_{k}$ signifies the proximity between $h^d_{k-1}$ and each $h^x_i$, and is calculated by:
\begin{equation}
    \alpha_{ki} = \frac{exp(f(h^d_{k-1}, h^x_i))}{\sum^I_{j=1}exp(f(h^d_{k-1}, h^x_j))}.
\end{equation}
$f$ calculates the unnormalized similarity between $h^d_{k-1}$ and $h^x_i$, usually in the following ways:
\begin{equation}
f(h^d_{k-1}, h^x_i)=\left\{
\begin{array}{ll}
(h^d_{k-1})^Th^x_i, & {\rm dot}\\
(h^d_{k-1})^TWh^x_i, & {\rm general}\\
v^t{\rm tanh}(W[h^d_{k-1}, h^x_i]), & {\rm concat}
\end{array} \right.
\end{equation}

\section{Proposed Method}
The framework of LIBS is illustrated in Figure \ref{fig:network}. 
Both the speech recognizer and the lip reader are based on the attention-based sequence-to-sequence architecture. 
For an input video, $\textbf{x}^v = [x^v_1, ..., x^v_J]$ represents its video frame sequence, $\textbf{y} = [y_1, ..., y_K]$ is the target character sequence.
The corresponding audio frame sequence is $\textbf{x}^a = [x^a_1, ..., x^a_I]$. 
A pre-trained speech recognizer reads in the audio frame sequence $\textbf{x}^a$, and outputs the predicted character sequence $\tilde{\textbf{y}} = [\tilde{y}_1, ..., \tilde{y}_L]$. 
It should be noted that the sentence predicted by speech recognizer is imperfect, and $L$ may not equal to $K$. 
At the same time, the encoder hidden states $\textbf{h}^a = [h^a_1, ..., h^a_I]$, sequence vector $s^a$, and context vectors $\textbf{c}^a = [c^a_1, ..., c^a_L]$ can also be obtained. 
They are used to guide the training of the lip reader.

The basic lip reader is trained to maximize conditional probability distribution $p(\textbf{y}|\textbf{x}^v)$, which equals to minimize the loss function:
\begin{equation}
    L_{base} = -\sum^K_{k=1} \log p(y_k|y_{<k}, \textbf{x}^v).
\end{equation}
The encoder hidden states, sequence vector and context vectors of the lip reader are denoted as $\textbf{h}^v = [h^v_1, ..., h^v_J]$, $s^v$, and $\textbf{c}^v = [c^v_1, ..., c^v_K]$, respectively.

The proposed method LIBS aims to minimize the loss function: 
\begin{equation}
    L = L_{base} + \lambda_1 L_{KD1} + \lambda_2 L_{KD2} + \lambda_3 L_{KD3} ,
\end{equation}
where $L_{KD1}$, $L_{KD2}$, and $L_{KD3}$ constitute the multi-granularity knowledge distillation, and work at sequence-level, context-level and frame-level respectively. 
$\lambda_1, \lambda_2$ and $\lambda_3$ are the corresponding balance weights. Details are described below. 

\subsection{Sequence-Level Knowledge Distillation}
As mentioned before, the sequence vector $s^x$ contains the semantic information of the input sequence. 
For a video frame sequence $\textbf{x}^v$ and its corresponding audio frame sequence $\textbf{x}^a$, their sequence vectors $s^a$ and $s^v$ should be the same, because they are different expressions of the same thing.

Therefore, the sequence-level knowledge distillation is denoted as :
\begin{equation}
    L_{KD1} = \left\| s^a - t(s^v)\right\|_2^2.
\end{equation}
$t$ is a simple transformation function (for example a linear or affine function), which embeds features into a space with the same dimension.

\subsection{Context-Level Knowledge Distillation}
When decoder predicting a character at a certain timestep, the attention mechanism uses context vector to summarize the input information that is most relevant to the current output. 
Therefore, if the lip reader and speech recognizer predict the same character at $j$-th timestep, the context vectors $c^v_j$ and $c^a_j$ should contain the same information. Naturally, the context-level knowledge distillation should push $c^v_j$ and $c^a_j$ to be the same. 

However, due to the imperfect speech-recognition predictions, it's possible that $\tilde{y}_j$ and $y_j$ may not be the same.
Simply making $c^v_j$ and $c^a_j$ similar would hinder the performance of lip reader.
This requires choosing the correct characters from the speech-recognition predictions, and using the corresponding context vectors for knowledge distillation. 
Besides, in current attention mechanism, the context vectors are built upon the RNN hidden state vectors, 
which act as representations of prefix substrings of the input sentences, given the sequential nature of RNN computation \cite{wu2018word}. 
Thus, even if there are same characters in the predicted sentence, 
their corresponding context vectors are different because of their different positions. 

Based on these findings, a Longest Common Subsequence (LCS) \footnote{\url{https://en.wikipedia.org/wiki/Longest_common_subsequence_problem}} based filtering method is proposed to refine the distilled features.  
LCS is used to compare two sequences. 
Common subsequences with same order in the two sequences are found, and the longest sequence is selected. 
The most important aspects of LCS are that the common subsequence is not necessary to be contiguous, and it retains the relative position information between characters.
Formally speaking, LCS computes the common subsequence between $\tilde{\textbf{y}} = [\tilde{y}_1, ..., \tilde{y}_L]$ and $\textbf{y} = [y_1, ..., y_K]$, 
and obtains the subscripts of the corresponding characters in $\tilde{\textbf{y}}$ and $\textbf{y}$:
\begin{equation}
\begin{aligned}
    {I}^a_1, ..., {I}^a_M, {I}^v_1, ..., {I}^v_M & = {\rm LCS}(\tilde{y}_1, ..., \tilde{y}_L, y_1, ..., y_K), \\[1mm]
    M & \leq \min(L, K),
\end{aligned}
\end{equation}
where ${I}^a_1, ..., {I}^a_M$ and ${I}^v_1, ..., {I}^v_M$ are the subscripts in the sentence predicted by speech recognizer and the ground truth sentence, respectively. 
Please refer to the supplementary material for details.
It's worth noting that when the sentence is Chinese, two characters are defined to be the same if they have the same Pinyin. 
Pinyin is the phonetic symbol of Chinese character, and homophones account for more than 85\% among all Chinese characters.

Context-level knowledge distillation only calculate on these common characters:
 \begin{equation}
     L_{KD2} = \frac{1}{M}\sum^M_{i=1} \left\| c^a_{{I}^a_i} - t(c^v_{{I}^v_i})\right\|_2^2.
 \end{equation}

\subsection{Frame-Level Knowledge Distillation} 
Furthermore, we hope that the speech recognizer can teach the lip reader more finely and explicitly. 
Specifically, knowledge is distilled at frame-level to enhance the discriminability of each video frame feature.

If the correspondence between video and audio is known, then it is sufficient to directly match the video frame feature with the corresponding audio feature. 
However, due to the different sampling rates, video sequence and audio sequence have inconsistent length. 
Besides, since blanks may appear at the beginning or end of the data, there is no guarantee that video and audio are strictly synchronized. 
Therefore, it is impossible to specify the correspondence artificially. 
This problem is solved by first learning the correspondence between video and audio, then performing the frame-level knowledge distillation.

As the hidden states of RNN providing higher-level semantics and are easier to correlated than the original input feature \cite{sterpu2018attention}, the alignment between audio and video is learned on the hidden states of the audio encoder and video encoder.
Formally speaking, for each audio hidden state $h^a_i$, the most similar video frame feature is calculated by a way similar to the attention mechanism:
\begin{equation}
    \tilde{h}^v_i = \sum^J_{j=1}\beta_{ji}h^v_j,
\end{equation}
$\beta_{ji}$ is the normalized similarity between $h^a_i$ and video encoder hidden states $h^v_j$:
\begin{equation}
    \beta_{ji} = \frac{exp((h^v_j)^TWh^a_i)}{\sum^J_{k=1}exp((h^v_k)^TWh^a_i)}.
\end{equation}
Since $\tilde{h}^v_i$ contains the most similar information to audio feature $h^a_i$ and the acoustic speech signals embody information complementary to the visual ones, making $\tilde{h}^v_i$ and $h^a_i$ the same enhances lip reader's capability to extract discriminant visual feature. 
Thus, the frame-level knowledge distillation is defined as:
\begin{equation}
    L_{KD3} =  \frac{1}{I}\sum^I_{i=1} \left\| h^a_i - \tilde{h}^v_i\right\|_2^2. 
\end{equation}

The audio and video modalities can have two-way interactions. However, in the preliminary experiment, we found that video attending audio leads to inferior performance. So, only audio attending video is chosen to perform the frame-level knowledge distillation.

\section{Experiments}

\subsection{Datasets}
\textbf{CMLR}\footnote{\url{https://www.vipazoo.cn/CMLR.html}}~\cite{zhao2019cascadelipreading}: it is currently the largest Chinese Mandarin lip reading dataset. 
It contains over 100,000 natural sentences from China Network Television website, 
including more than 3,000 Chinese characters and 20,000 phrases. \\
\textbf{LRS2}\footnote{\url{http://www.robots.ox.ac.uk/~vgg/data/lip_reading/lrs2.html}}~\cite{afouras2018deep}: it contains more than 45,000 spoken sentences from BBC television. 
LRS2 is divided into development (train/val) and test sets according to the broadcast date. 
The dataset has a "\textit{pre-train}" set that contains sentences annotated with the alignment boundaries of every word. 

We follow the provided dataset partition in experiments. 

\subsection{Evaluation Metrics}
For experiments on LRS2 dataset, we report the Character Error Rate (CER), Word Error Rate (WER) and BLEU \cite{papineni2002bleu}. 
The CER and WER are defined as $ErrorRate = (S + D + I) / N$, 
where $S$ is the number of substitutions, 
$D$ is the number of deletions, 
$I$ is the number of insertions to get from the reference to the hypothesis and 
$N$ is the number of characters (words) in the reference. 
BLEU is a modified form of n-gram precision to compare a candidate sentence to one or more reference sentences. 
Here, the unigram BLEU is used. 
For experiments on CMLR dataset, only CER and BLEU are reported, 
since the Chinese sentence is presented as a continuous string of characters without demarcation of word boundaries.

\subsection{Training Strategy}
Same as \cite{chung2017lipWild}, curriculum learning is employed to accelerate training and reduce over-fitting. 
Since the training sets of CMLR and LRS2 are not annotated with the word boundaries, 
the sentences are grouped into subsets according to the length.
We start training on short sentences and 
then make the sequence length grow as the network trains.
Scheduled sampling \cite{bengio2015scheduled} is used to eliminate the discrepancy between training and inference. The sampling rate from the previous output is selected from 0.7 to 1 for CMLR dataset, and from 0 to 0.25 for LRS2 dataset.
For fair comparisons, decoding is performed with beam search of width 1 for CMLR and 4 for LRS2, in a similar way to \cite{chan2016listen}.

However, preliminary experimental results show that the sequence-to-sequence based model is hard to achieve reasonable results on the LRS2 dataset. 
This is because even the shortest English sentence contains 14 characters, 
which is still difficult for the decoder to extract relevant information from all input steps at the beginning of the training. 
Therefore, a pre-training stage is added for LRS2 dataset as in \cite{afouras2018deep}. 
When pre-training, the CNN pre-trained on word excerpts from the MV-LRS \cite{chung2017lip} dataset is used to extract visual features for the \textit{pre-train} set. 
The lip reader is trained on these frozen visual features.
Pre-training starts with a single word, then gradually increases to a maximum length of 16 words. After that, the model is trained end-to-end on the training set.

\subsection{Implementation Details}

\subsection{Lip Reader}
\textbf{CMLR}: 
The input images are 64 $\times$ 128 in dimension. 
VGG-M model\cite{chatfield2014return} is used to extract visual features. 
Lip frames are transformed into gray-scale, and the VGG-M network takes every 5 lip frames as an input, moving 2 frames at each timestep. 
We use a two-layer bi-directional GRU \cite{cho2014learning} with a cell size of 256 for the encoder and a two-layer uni-directional GRU with a cell size of 512 for the decoder. 
For character vocabulary, characters that appear more than 20 times are kept. [sos], [eos] and [pad] are also included. 
The final vocabulary size is 1,779. The initial learning rate was 0.0003 and decreased by 50\% every time the training error did not improve for 4 epochs. 
Warm-up \cite{he2016deep} is used to prevent over-fitting.\\
\textbf{LRS2}: 
The input images are 112 $\times$ 112 pixels covering the region around the mouth. 
The CNN used to extract visual features is based on \cite{stafylakis2017combining}, with a filter width of 5 frames in 3D convolutions. 
The encoder contains 3 layers of bi-directional LSTM \cite{hochreiter1997long} with a cell size of 256,
and the decoder contains 3 layers of uni-directional LSTM with a cell size of 512. 
The output size of lip reader is 29, containing 26 letters and tokens for [sos], [eos], [pad].
The initial learning rate was 0.0008 for pre-training, 0.0001 for training, and decreased by 50\% every time the training error did not improve for 3 epochs. 

The balance weights used in both datasets are shown in Table \ref{table:balance_weights}. The values are obtained by conducting a grid search. 

\begin{table}
\caption {The balance weights employed in CMLR and LRS2 datasets.}
\label{table:balance_weights}
\centering
\small
\begin{tabular}{p{0.33\columnwidth} | p{0.15\columnwidth}<{\centering} | p{0.15\columnwidth}<{\centering} | p{0.15\columnwidth}<{\centering}}
  \hline \textbf{Dataset} & \textbf{$\lambda_1$} & \textbf{$\lambda_2$} & \textbf{$\lambda_3$} \\ \hline 
  \textbf{CMLR}        &   10  &    40    &    10   \\ \hline
  \textbf{LRS2}    &    2  &    10    &    10  \\ \hline
\end{tabular}
\end{table}

\subsection{Speech Recognizer}
The datasets used to train speech recognizers are the audio of the CMLR and LRS2 datasets, plus additional speech data: aishell \cite{bu2017aishell} for CMLR, and LibriSpeech \cite{panayotov2015librispeech} for LRS2.
The 240-dimensional fbank feature is used as the speech feature, sampled at 16kHz and calculated over 25ms windows with a step size 10ms.  
For LRS2 dataset, the speech recognizer and lip reader have the same architecture. 
For CMLR dataset, specifically, three different speech recognizer architectures are considered to verify the generalization of LIBS.\\
\textbf{Teacher 1}: It contains 2 layers of bi-directional GRU for encoder with a cell size of 256, 2 layers of uni-directional GRU for decoder with a cell size 512. In other words, it has the same architecture as lip reader.\\
\textbf{Teacher 2}: The cell size of both encoder and decoder is 512. Others remain the same as Teacher 1. \\
\textbf{Teacher 3}: The encoder contains 3 layers of pyramid bi-directional GRU \cite{chan2016listen}. Others remain the same as Teacher 1.\\
It's worth noting that Teacher 2 and the lip reader have different feature dimensions,
and Teacher 3 reduces the audio time resolution by 8 times.

\subsection{Experimental Results}
\subsubsection{Effect of different teacher models.}
\begin{table}
\caption {The performance of LIBS when using different teacher models on the CMLR dataset.}
\label{table:comparison_teachers}
\centering
\small
\begin{tabular}{p{0.33\columnwidth} | p{0.25\columnwidth}<{\centering} | p{0.25\columnwidth}<{\centering}}
  \hline \textbf{Model} & \textbf{BLEU} & \textbf{CER} \\ \hline 
    \hline
    WAS                 & 64.13     &   38.93\% \\ \hline
    \hline
    Teacher 1           & 90.36     &   9.83\%   \\ 
    \textbf{LIBS}       & \textbf{69.99} & \textbf{31.27\%}  \\ 
    \hline
    Teacher 2           & 90.95 &   9.23\%       \\ 
    LIBS                & 66.66 &   34.94\%     \\ 
    \hline
    Teacher 3           & 87.73  &   12.40\%     \\ 
    LIBS                & 66.58  &   34.76\%     \\ \hline
\end{tabular}
\end{table}

To evaluate the generalization of the proposed multi-granularity knowledge distillation method, we compare the effects of LIBS on the CMLR dataset under different teacher models. 
Since WAS \cite{chung2017lipWild} and the baseline lip reader (trained without knowledge distillation) have the same sequence-to-sequence architecture, 
WAS is trained using the same training strategy as LIBS, 
and is used interchangeably with \textit{baseline} in the paper.
As can be seen from Table \ref{table:comparison_teachers}, LIBS substantially exceeds the baseline under different teacher model architectures.  
It is worth noting that although the performance of Teacher 2 is better than that of Teacher 1, the corresponding student network is not. 
This is because the feature dimensions of Teacher 2 speech recognizer and lip reader are different. 
This implies that distill knowledge directly in the same dimensional feature space can achieve better results.
In the following experiments, we analyze the lip reader learned from Teacher 1 on the CMLR dataset.

\subsubsection{Effect of the multi-granularity knowledge distillation.}
\begin{table}
\caption {Effect of the proposed multi-granularity knowledge distillation.}
\label{table:comparison_ablation}
\centering
\small
\begin{tabular}{p{0.38\columnwidth} | p{0.13\columnwidth}<{\centering} | p{0.15\columnwidth}<{\centering} | p{0.15\columnwidth}<{\centering}}
  \hline \textbf{Methods} & \textbf{BLEU} & \textbf{CER} & \textbf{WER}  \\ \hline 
    \hline
    \multicolumn{4}{c}{\textbf{CMLR}} \\
    \hline
    WAS                        & 64.13 & 38.93\% &   -  \\ 
    \hline
    WAS $ + L_{KD1}$           & 67.23 & 34.42\% &   -  \\ 
    WAS $ + L_{KD2}$           & 68.24 & 33.17\% &   -  \\ 
    WAS $ + L_{KD3}$           & 66.31 & 35.30\% &   -  \\ 
    WAS $ + L_{KD1} + L_{KD2}$ & 68.53 & 32.95\% &   -  \\ \hline
    LIBS                        & \textbf{69.99} & \textbf{31.27\%} &   -  \\ \hline
    \hline
    \multicolumn{4}{c}{\textbf{LRS2}} \\
    \hline
    WAS                        & 39.72 & 48.28\% & 68.19\%   \\ 
    \hline
    WAS $ + L_{KD1}$           & 41.00 & 46.04\% & 66.59\%  \\ 
    WAS $ + L_{KD2}$           & 41.23 & 46.01\% & 66.31\%  \\ 
    WAS $ + L_{KD3}$           & 41.18 & 46.91\% & 66.65\%     \\ 
    WAS $ + L_{KD1} + L_{KD2}$ & 41.55 & 45.97\% & 65.93\%  \\ \hline
    LIBS                        & \textbf{41.91} & \textbf{45.53\%} & \textbf{65.29\%} \\ \hline
\end{tabular}
\end{table}

Table \ref{table:comparison_ablation} shows 
the effect of the multi-granularity knowledge distillation on CMLR and LRS2 datasets.
Comparing WAS, WAS $ + L_{KD1}$, WAS $ + L_{KD1} + L_{KD2}$ and LIBS, 
all metrics are increasing along with adding different granularity of knowledge distillation. 
The increasing results show that each granularity of knowledge distillation is able to contribute to the performance of LIBS. 
However, the smaller and smaller extent of the increase does not indicate that 
the sequence-level knowledge distillation has greater influence than the frame-level knowledge distillation.
When only one granularity of knowledge distillation is added, 
WAS $ + L_{KD2}$ shows the best performance. 
This is due to the design that the context-level knowledge distillation is directly acting on the features used to predict characters. 

On the CMLR dataset, LIBS exceeds WAS by a margin of 7.66\% in CER. However, the margin is not that large on the LRS2 dataset, only 2.75\%. 
This may be caused by the differences in the training strategy. On LRS2 dataset, CNN is first pre-trained on the MV-LRS dataset. 
Pre-training gives CNN a good initial value so that better video frame feature can be extracted during the training process. 
To verify this, we compare WAS and LIBS trained without the pre-training stage. 
The CER of WAS and LIBS are 67.64\% and 62.91\% respectively, with a larger margin of 4.73\%. 
This confirms the hypothesis that LIBS can help to extract more effective visual features.

\subsubsection{Effect of different amount of training data.}
\begin{table}
\caption {The performance of LIBS when trained with different amount of training data on the CMLR dataset.}
\label{table:comparison_limited}
\centering
\small
\begin{tabular}{c|c|c|c|c}
\hline
\begin{tabular}[c]{@{}l@{}}
\textbf{Percentage of} \\ \textbf{Training Data}\end{tabular} & \textbf{Metrics} & \textbf{WAS} & \textbf{LIBS} & \textbf{Improv} \\ 
\hline 
\hline
\multirow{2}{*}{100\%}  & CER  & 38.93\% & 31.27\%  & 7.66\%  $\downarrow$    \\ 
\cline{2-5} & BLEU & 64.13   & 69.99    & 5.86 $\uparrow$ \\ \hline
                        
\hline
\multirow{2}{*}{20\%}   & CER  & 60.13\% &  50.50\% & \textbf{9.63\%} $\downarrow$ \\ 
\cline{2-5} & BLEU & 42.69   &  50.65   & \textbf{7.96} $\uparrow$  \\ \hline

\end{tabular}
\end{table}
Compared with lip video data, the speech data is easier to collect. 
We evaluate the effect of LIBS in the case of limited lip video data on CMLR dataset.
As mentioned before, the sentences are grouped into subsets according to the length, and only the first subset is used to train the lip reader. 
The first subset is about 20\% of the full training set, 
which contains 27,262 sentences, and the number of characters in each sentence does not exceed 11. 
It can be seen from the Table \ref{table:comparison_limited}, 
when the training data is limited, LIBS tends to yield an even greater performance gain:
the improvement on CER increases from 7.66\% to 9.63\%, and from 5.86 to 7.96 on BLEU.

\subsubsection{Comparison with state-of-the-art methods.}
\begin{table}
\caption {Performance comparison with other existing frameworks on the CMLR and LRS2 datasets. }
\label{table:comparison_STOA}
\centering
\small
\begin{tabular}{p{0.21\columnwidth} | p{0.2\columnwidth}<{\centering} | p{0.2\columnwidth}<{\centering} | p{0.2\columnwidth}<{\centering}}
    \hline 
    \textbf{Methods}    &  \textbf{BLEU} & \textbf{CER} & \textbf{WER}  \\ \hline 
    \hline
    \multicolumn{4}{c}{\textbf{CMLR}} \\
    \hline
    WAS             &      64.13    &      38.93\%      &   -  \\ \hline
    CSSMCM          &        -      &      32.48\%      &   -  \\ \hline
    LIBS            & \textbf{69.99}& \textbf{31.27\%}  &   -  \\ \hline
    \hline
    \multicolumn{4}{c}{\textbf{LRS2}} \\
    \hline
    WAS             &   39.72   &       48.28\%     & 68.19\%   \\ \hline
    TM-seq2seq      &   -       &       -     & \textbf{49.8\%}  \\ \hline
    CTC/Attention   &   -       &    \textbf{42.1\%}  & 63.5\% \\ \hline
    LIBS   & 41.91 & 45.53\% & 65.29\% \\ \hline
\end{tabular}
\end{table}

\begin{figure*}[ht]
\centering
\subcaptionbox{WAS} % 1st
{
\includegraphics[width=0.5\columnwidth]{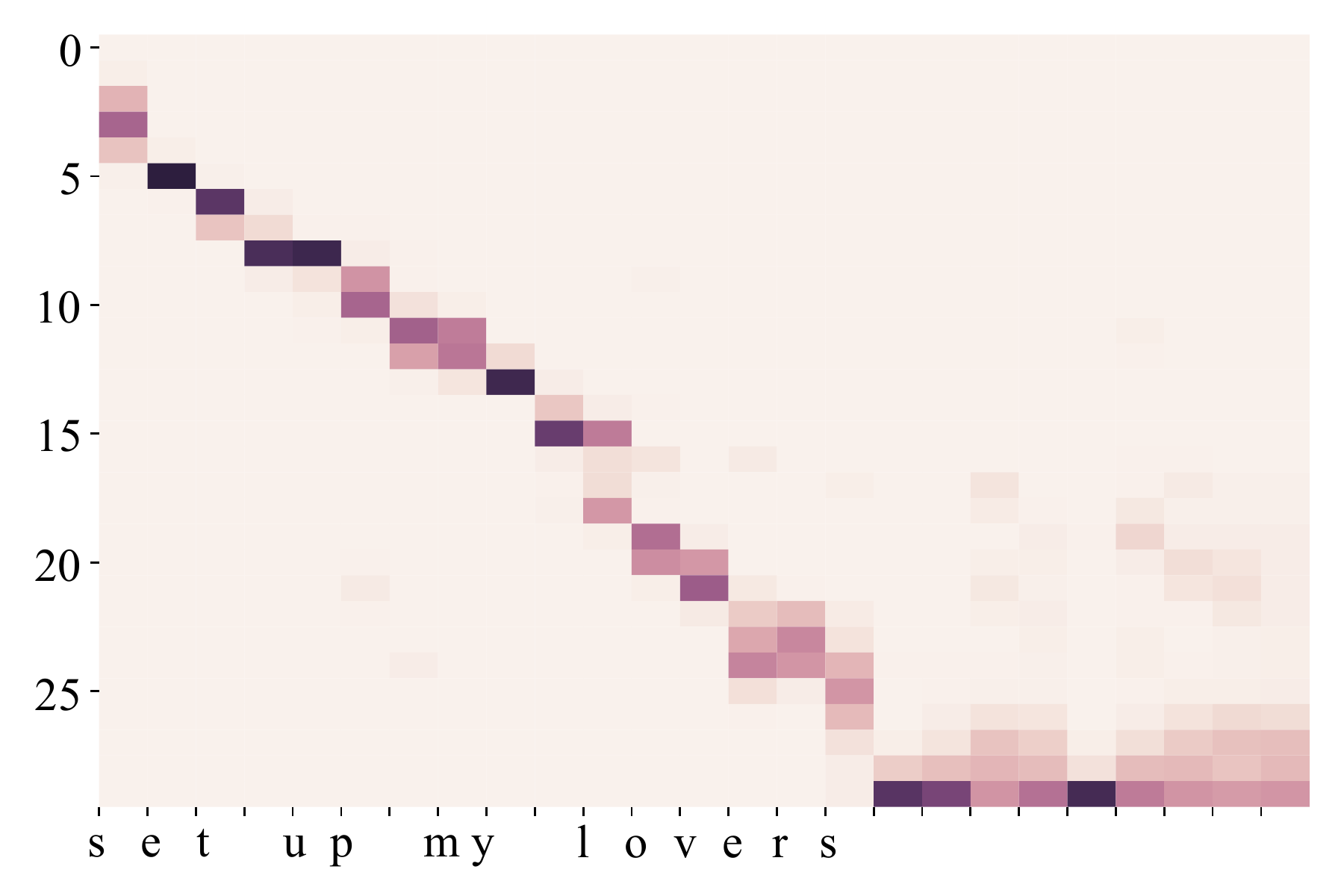}
}
\subcaptionbox{WAS $ + L_{KD1}$} % 2nd
{
\includegraphics[width=0.5\columnwidth]{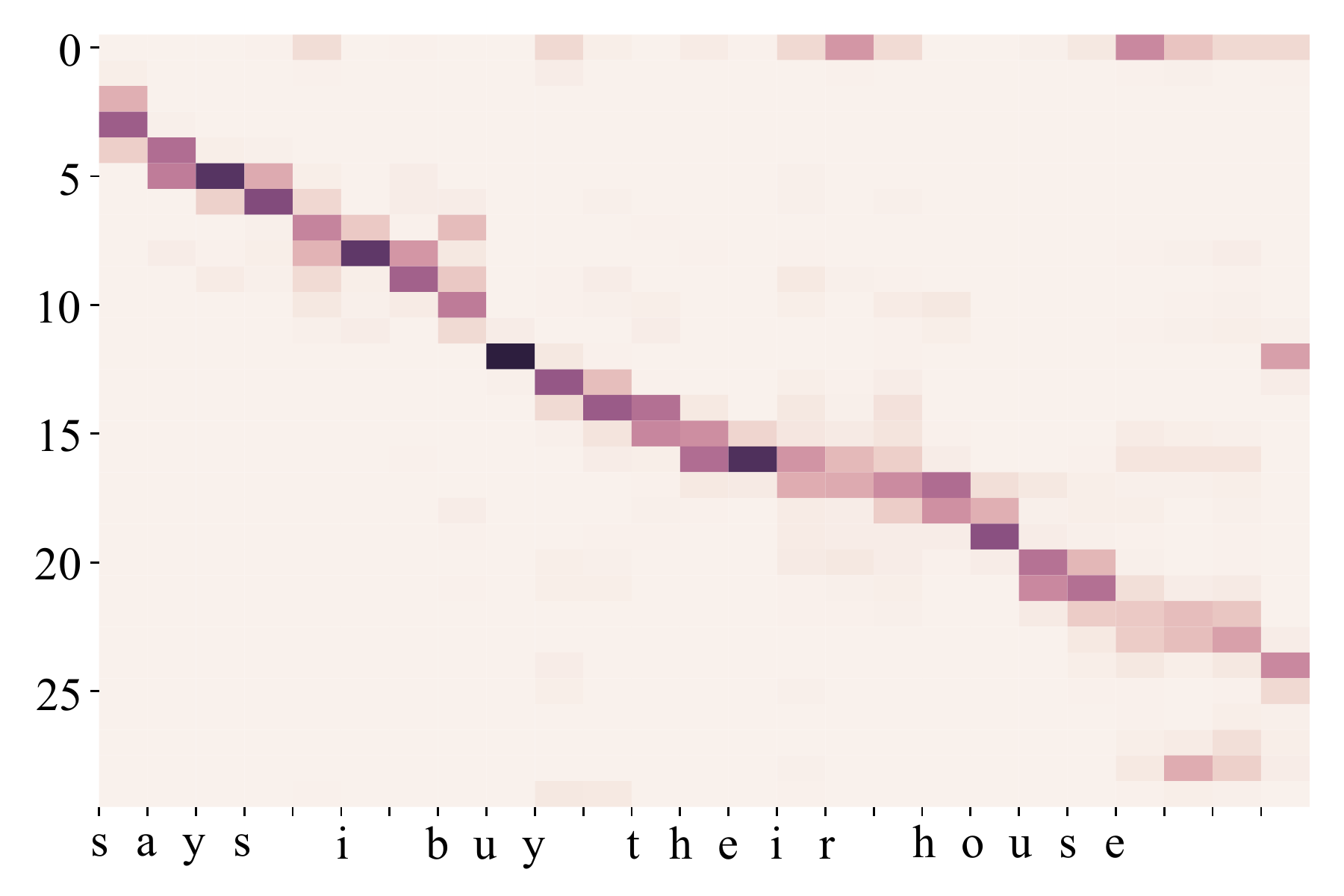}
}
\subcaptionbox{WAS $ + L_{KD1} + L_{KD2}$} % 3rd
{
\includegraphics[width=0.5\columnwidth]{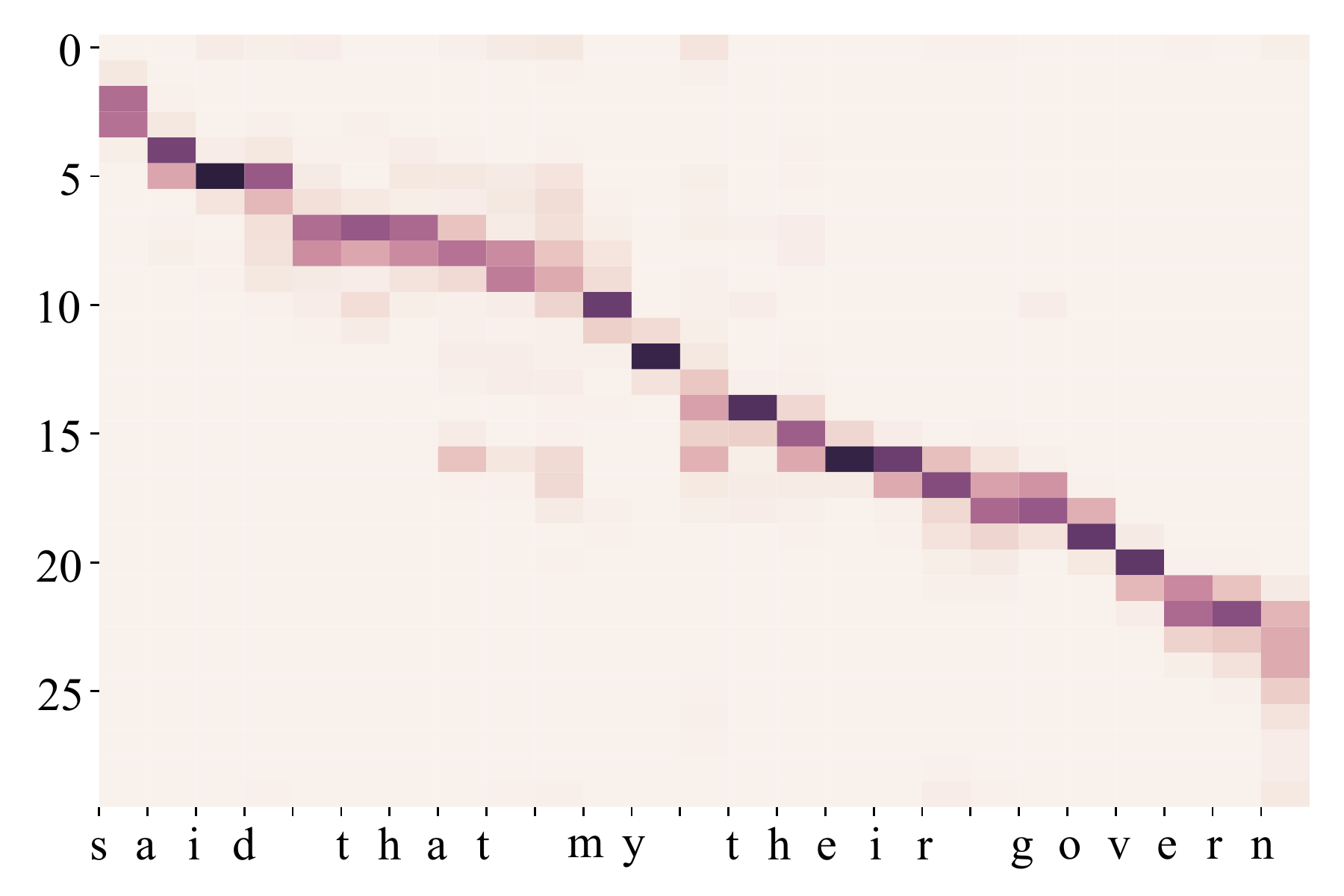}
}
\subcaptionbox{LIBS} % 4th
{
\includegraphics[width=0.5\columnwidth]{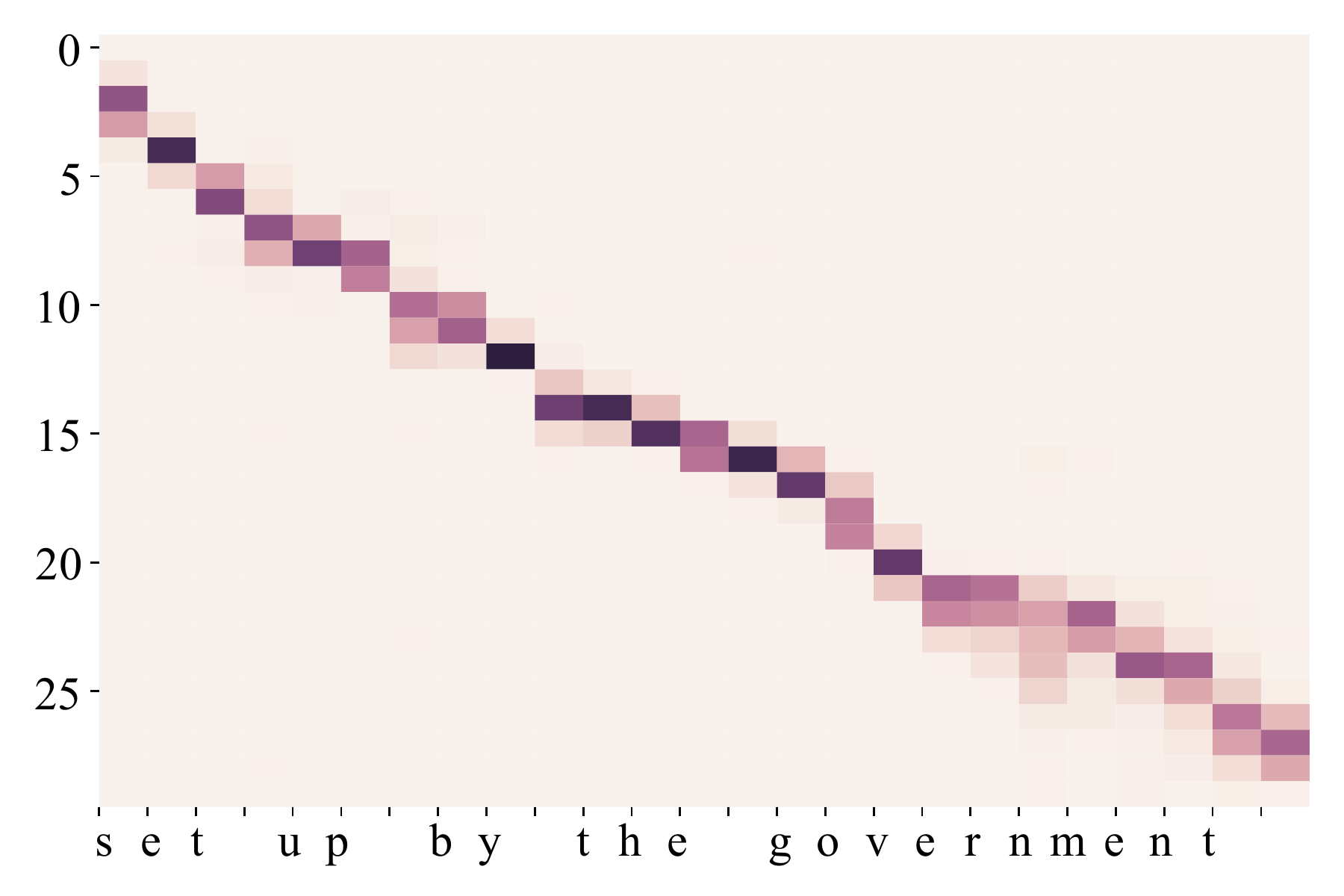}
}
 
\caption{Alignment between the video frames and the predicted characters with different levels of the proposed multi-granularity knowledge distillation. The vertical axis represents the video frames and the horizontal axis represents the predicted characters. The ground truth sentence is \textit{set up by the government}.} %  %大图名称
\label{fig:attention}
\end{figure*}

% \begin{figure*}[ht]
% \centering
% \subfigure[WAS] % 1st
% {
% 	\centering 
% 	\includegraphics[width=0.5\columnwidth]{baseline.pdf}
% }
% \subfigure[WAS $ + L_{KD1}$] % 2nd
% {
% 	\centering 
% 	\includegraphics[width=0.5\columnwidth]{s2.pdf}
% }
% \subfigure[WAS $ + L_{KD1} + L_{KD2}$] %3rd
% {
% 	\centering 
% 	\includegraphics[width=0.5\columnwidth]{s2_f10.pdf}
% }
% \subfigure[LIBS] % 4th
% {
% 	\centering      %子图居中
% 	\includegraphics[width=0.5\columnwidth]{s2_f10_a10.pdf}
% }
 
% \caption{Alignment between the video frames and the predicted characters with different levels of the proposed multi-granularity knowledge distillation. The vertical axis represents the video frames and the horizontal axis represents the predicted characters. The ground truth sentence is \textit{set up by the government}.} %  %大图名称
% \label{fig:attention}
% \end{figure*}

\begin{figure*}[ht]
\centering
\setlength{\tabcolsep}{1pt}
\subcaptionbox{WAS} % 1st
{
	\centering 
	\begin{tabular}{ ccccccccccccc }
    \includegraphics[width=0.15\columnwidth]{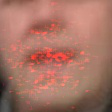} & \includegraphics[width=0.15\columnwidth]{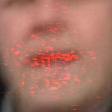} &
    \includegraphics[width=0.15\columnwidth]{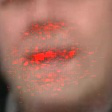} & \includegraphics[width=0.15\columnwidth]{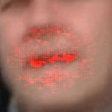} & \includegraphics[width=0.15\columnwidth]{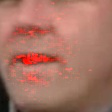} & \includegraphics[width=0.15\columnwidth]{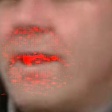} & \includegraphics[width=0.15\columnwidth]{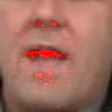} & \includegraphics[width=0.15\columnwidth]{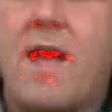} & \includegraphics[width=0.15\columnwidth]{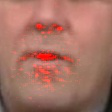} & \includegraphics[width=0.15\columnwidth]{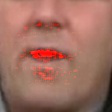} & \includegraphics[width=0.15\columnwidth]{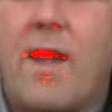} & \includegraphics[width=0.15\columnwidth]{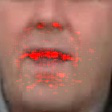} &
    \includegraphics[width=0.15\columnwidth]{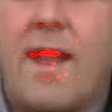}\\ 
    \end{tabular}
}

\subcaptionbox{LIBS} % 2nd
{
	\centering 
    \begin{tabular}{ ccccccccccccc } 
    \includegraphics[width=0.15\columnwidth]{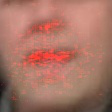} & \includegraphics[width=0.15\columnwidth]{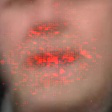} &
    \includegraphics[width=0.15\columnwidth]{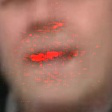} & \includegraphics[width=0.15\columnwidth]{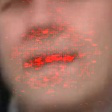} & \includegraphics[width=0.15\columnwidth]{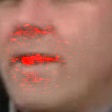} & \includegraphics[width=0.15\columnwidth]{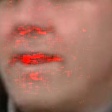} & \includegraphics[width=0.15\columnwidth]{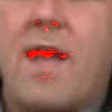} & \includegraphics[width=0.15\columnwidth]{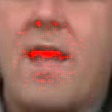} & \includegraphics[width=0.15\columnwidth]{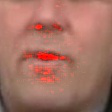} & \includegraphics[width=0.15\columnwidth]{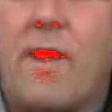} & \includegraphics[width=0.15\columnwidth]{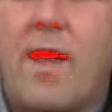} & \includegraphics[width=0.15\columnwidth]{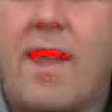} &
    \includegraphics[width=0.15\columnwidth]{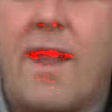}\\ 
    \end{tabular}
}

\caption{Saliency maps for WAS and LIBS. The places where the lip reader has learned to attend are highlighted in red .} %大图名称
\label{fig:saliency_map}
\end{figure*}

Table \ref{table:comparison_STOA} shows the experimental results compared with other frameworks: WAS \cite{chung2017lipWild}, CSSMCM \cite{zhao2019cascadelipreading}, TM-seq2seq \cite{afouras2018deep} and CTC/attention \cite{petridis2018audio}.
TM-seq2seq achieves the lowest WER on the LRS2 dataset due to its transformer self-attention architecture \cite{vaswani2017attention}. Since LIBS is designed for the sequence-to-sequence architecture, performance may be improved by replacing RNN with transformer self-attention block. Note that, despite the excellent performance of CSSMCM, which is designed for Chinese Mandarin lip reading, LIBS still exceeds it by a margin of 1.21\% in CER.

\subsection{Visualization}
\subsubsection{Attention visualization.}
The attention mechanism generates explicit alignment between the input video frames and the generated character outputs. 
Since the correspondence between the input video frames and the generated character outputs is monotonous in time, whether alignment has a diagonal trend is a reflection of the performance of the model \cite{wang2017tacotron}. 
Figure \ref{fig:attention} visualizes the alignment of the video frames and the corresponding outputs with different granularities of knowledge distillation on the test set of LRS2 dataset. 
Comparing Figure \ref{fig:attention}(a) with Figure \ref{fig:attention}(b), adding sequence-level knowledge distillation improves the quality of the end part of the generated sentence.
This indicates that the lip reader enhances its understanding of the semantic information of the whole sentence. 
Adding context-level knowledge distillation (Figure \ref{fig:attention}(c)) allows the attention at each decoder step to be concentrated around the corresponding video frames, reducing the focus on unrelated frames.
This also makes the predicted characters more accurate.
Finally, the frame-level knowledge distillation (Figure \ref{fig:attention}(d)) further improves the discriminability of the video frame features, making the attention more focused.
The quality and the comprehensibility of the generated sentence is increased along with adding different levels of knowledge distillation.

\subsubsection{Saliency maps.}
Saliency visualization technique is employed to verify that LIBS enhances lip reader's ability to extract discriminant visual features, 
by showing areas in the video frames the model concentrated most when predicting.
Figure \ref{fig:saliency_map} shows saliency visualisations for the baseline model and LIBS respectively, based on \cite{smilkov2017smoothgrad}.
Both the baseline model and LIBS can correctly focus on the area around the mouth, 
but the salient regions for baseline model are more scattered compared with LIBS.

\section{Conclusion}
In this paper, we propose LIBS, 
an innovative and effective 
approach to training lip reading
by learning from a pre-trained speech recognizer. 
LIBS distills speech-recognizer 
knowledge of multiple granularities,
from sequence-, context-, and frame-level, 
to guide the learning of the lip reader.
Specifically, this is achieved by 
introducing a novel
filtering strategy to refine the features from
the speech recognizer,
and by adopting a cross-modal alignment-based 
method for frame-level knowledge distillation
to account for the sampling-rate inconsistencies
between the two sequences. 
Experimental results demonstrate that the proposed
LIBS yields a considerable improvement over the state of the art,
especially when the training samples are limited. 
In our future work, we look forward to 
adopting the same framework to other modality pairs such as
speech and sign language. 

\section{Acknowledgements}
This work is supported by  National Key Research and Development Program (2016YFB1200203) , National Natural Science Foundation of China (61976186),  Key Research and Development Program of Zhejiang Province (2018C01004), and the Major Scientifc Research Project of Zhejiang Lab (No. 2019KD0AC01) .

\bibliographystyle{aaai}
\bibliography{ref} 

\end{document}